\documentclass[10pt, letterpaper]{article}
\usepackage[utf8]{inputenc}
\usepackage{graphicx}
\usepackage{amsmath}
\usepackage{amsthm}
\usepackage{amsfonts}
\usepackage[a4paper, total={6in, 8in}]{geometry}

\usepackage{natbib}
\usepackage{authblk}

\begin{document}
\title{Action Forecasting with Feature-wise Self-Attention}
\author{Yan Bin Ng}
\author{Basura Fernando}
\affil{IHPC, A*STAR, Singapore.}

\date{}

\maketitle
\begin{abstract}
We present a new architecture for human action forecasting from videos.
A temporal recurrent encoder captures temporal information of input videos while a self-attention model is used to attend on relevant feature dimensions of the input space.
To handle temporal variations in observed video data, a feature masking techniques is employed.
We classify observed actions accurately using an auxiliary classifier which helps to understand what has happened so far.
Then the decoder generates actions for the future based on the output of the recurrent encoder and the self-attention model.
Experimentally, we validate each component of our architecture where we see that the impact of self-attention to identify relevant feature dimensions, temporal masking, and observed auxiliary classifier.
We evaluate our method on two standard action forecasting benchmarks and obtain state-of-the-art results.
\end{abstract}

\section{Introduction}
\begin{figure}[t]
	\includegraphics[width=.99\columnwidth]{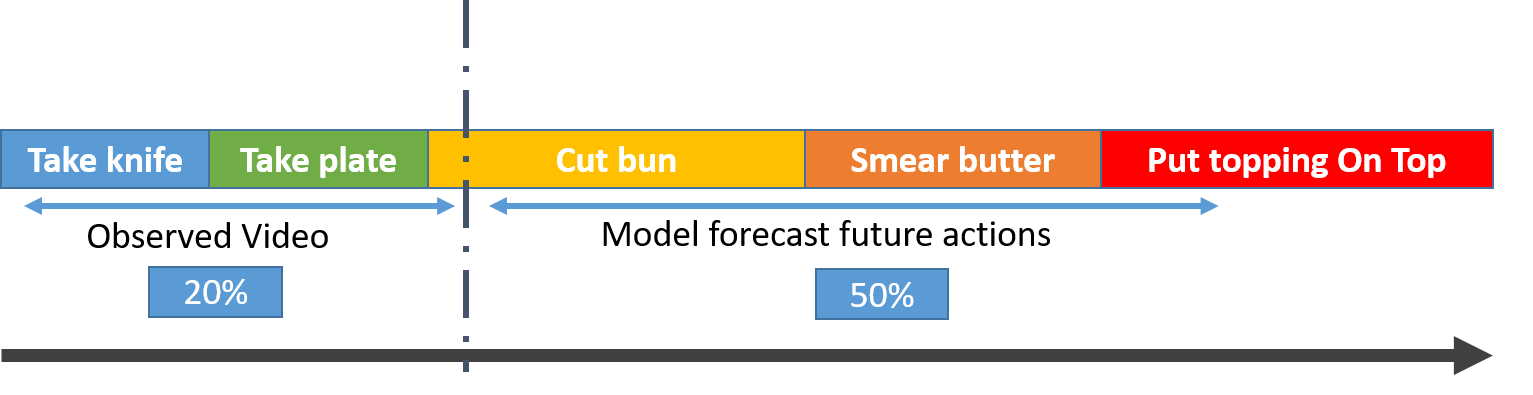}
	\caption{Action forecasting illustration studied in this paper. Given a long video of several minutes consisting of many actions, we observe p\% of the video and forecast actions for q\% of future frames. This task is different from early action recognition or action anticipation.}
\label{fig.illus}
\end{figure}

Video action forecasting aims at predicting future sequence of human actions from a partial observation of a video~\citep{AbuFarha2018}. 
As shown in figure~\ref{fig.illus}, the model observes a certain percentage of the activity (of making a sandwich) and then predicts future actions for a certain percentage of the unobserved video. 
This problem is more challenging than video action recognition~\citep{Simonyan2014,Wang2013}, early action recognition~\citep{Ryoo2011,Gammulle2019} and action anticipation~\citep{Lan2014,Furnari2019}. 
Action recognition aims to predict an action from a video after the action is fully observed.
In early action recognition, the objective is to recognize an ongoing action as early as possible before it finishes.
Action anticipation aims to predict a single future action that is going to happen in the immediate future before the action starts.
Action forecasting aims at predicting a sequence of actions in the chronological order that they are going to happen in each future frame~\citep{AbuFarha2018,ke2019time}, which is the objective of this work.
While it is a very important problem for tasks such as human robotic collaboration and human behavior understating, action forecasting is gaining some popularity in recent years~\citep{AbuFarha2018,ke2019time}.

Action forecasting models should solve several challenging problems. First, the model should know the intention of the human and the actions that human has performed so far to accomplish the goal. However, the model sees the activity only partially. From this partial observation, the model should aim to guess what the human might do in the future. Action forecasting models should have a good understanding of the activity being performed and the observed actions. 
Secondly, each human might perform the same activity in different ways with different speeds. Therefore, there is a large variation in the observed video sequence for the same activity. 
For example the activity of making a sandwich could be initiated by taking the bread or by taking the plate or a knife. Some may smear butter on both slices and some may not. There is a large amount of variation in the steps that one may follow even for a simple task such as making sandwich.

In this paper we solve the above two challenges using three techniques.
First, our model extracts a spatial temporal feature sequence from the observed video using inflated 3D convolutional networks~\citep{Carreira2017}.
Then we propose a novel encoder decoder architecture to forecast future actions for future unobserved frames.
Our temporal recurrent encoder consists of a GRU model and a self-attention-based model to weight the most relevant temporal feature dimensions of observed video sequence. This allows us to better model the temporal information. GRU model captures the temporal evolution of features while the feature-based self-attention identifies the relevant feature dimensions of the signal that are important for action understanding and forecasting. 
Typically, self-attention modules are applied in the temporal axis to identify the most relevant features of the sequence. 
In our case, we use the self-attention mechanism in an orthogonal direction to identify the most relevant feature dimensions of the temporal signal.
Experimentally, we demonstrate the impact of this design choice.

We concatenate the output of the GRU and feature-based self-attention model to predict observed actions. This allows our model to better understand what has happened already in the observed video sequence. 
The GRU decoder then takes the concatenated summary representation of the input to generate actions for the future. To model temporal variations in the observed signal, we employ a new technique called feature masking. Feature masking randomly drops features of the observed video sequence making temporal variations more explicit during the training of the model. Feature masking is somewhat similar to dropout, however it differs from dropout due to two reasons. First, feature masking masks out entire features and therefore zero features are not presented to the network. 
While dropout removes the over-reliance on some neurons in a network, feature masking acts as temporal data augmentation to model temporal variations. Specifically, this simple technique helps to boost action forecasting performance.

In summary, our main contribution is a new architecture that consists of three key components to boost action forecasting performance on two standard action forecasting benchmarks, the 50Salads and the Breakfast dataset.

\section{Related work}
The first work on action forecasting is presented in year 2018 by \citeauthor{AbuFarha2018}. This pioneering work proposed two models to predict future actions from partially observed video. In their experimentation, they observe p\% of the activity and predict actions for q\% of the video. Input to their methods are the one-hot vector representation of class information of observed frames ($\left< y_1, \cdots y_{n_{obs}} \right>$ ). Their Hidden-Markov-Model-based  Recurrent Neural Network (RNN) first predicts the actions of the observed frames, which is then fed back to another RNN to predict future actions. In their second model, the CNN takes the one hot score matrix of the observed frames, then directly predicts the future action scores of the future frames.
In contrast to these models, our approach is fundamentally different. 
First, our method is feature-based and we do not directly rely on predicted actions of the observed videos to generate future actions. We encode two types of information from the observed frame feature sequence using GRU encoder and a self-attention-based model to identify relevant feature dimensions.
We use an auxiliary classifier to predict actions of the observed frames which aims to make sure that the temporal encoding obtained from the GRU encoder contains observed class information. We use a GRU decoder to decode the temporal summary of input features to predict actions of the future.

To overcome the iterative nature of RNN-based action forecasting model~\citep{AbuFarha2018}, the work in~\citep{ke2019time} proposed to condition on time. In particular, they directly aim to predict an action of the future after $t$ amount of time. To do that, they obtain a time representation for the anticipation time $t$ by a simple encoding and then concatenate with the observed action score matrix to obtain a joint representation, which is then processed by a multi-scale attention based convolution. The output of the convolution is then used to predict that action after $t$ amount of time. In contrast to \citep{ke2019time,AbuFarha2018}, we do not rely on action scores of the observed data to predict future action. We make use of the temporal representation of observed frames to make an inference on future actions. Secondly, our attention mechanism is applied over the feature dimensions to identify temporally relevant feature dimensions and we use a GRU and self-attention to obtain a rich temporal representation that is useful for future action forecasting. We enrich the obtained temporal representation by an auxiliary classifier, so that the final representation presented to the GRU decoder entails necessary information to understand the intention and overall activity of the human. Weakly supervised action forecasting method is also proposed in~\citep{Ng2020}.

Action forecasting is a relatively young topic and its closest sister problems are action anticipation~\citep{Lan2014,Furnari2019,Miech2019,Fernando2021} and early action recognition~\citep{Ryoo2011,Gammulle2019,Zhao2019,shi2018action}.
As we aim to predict far into the future (as long as 5 minutes in some videos), our action forecasting is a more challenging problem than these methods.

\section{Method}
\subsection{Problem}
Given an input video in which a human is performing an activity, our model aims to predict the future unseen action sequence while observing only an initial part of the video containing an initial action sequence.

Formally, given a RGB video $\mathbf{F} = \left< \mathbf{f}_1, \mathbf{f}_2, \cdots, \mathbf{f}_n \right >$ where $n$ is the number of video frames, we observe only an initial part of the video $\mathbf{F_{obs}} = \left< \mathbf{f}_1, \mathbf{f}_2, \cdots, \mathbf{f}_{n_{obs}} \right >$ and aim to predict the actions for a future portion of the video $\mathbf{Y} = \left< \mathbf{y}_1, \mathbf{y}_2, \cdots, \mathbf{y}_{n_f} \right>$ where $n_{obs}$ is the number of observed video frames and $n_f$ is the number of frames which we want to predict future actions. 
The observed video $F_{obs}$ and the future sequence $\mathbf{Y}$ may contain more than one action.
Given a video, we observe $p\%$ of the frames and predict actions for $q\%$ of the future frames of the video as done in prior work~\cite{AbuFarha2018,ke2019time}.

\begin{figure}[ht]
	\includegraphics[width=.99\columnwidth]{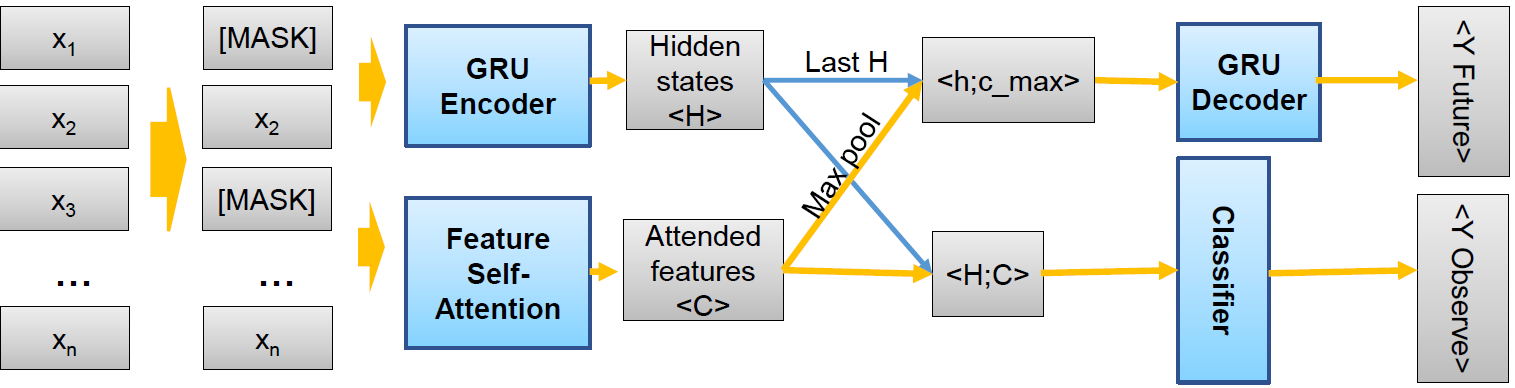}
	\caption{A visual illustration of our model architecture. 
It consists of a GRU encoder, a feature-wise self-attention transformer and a GRU decoder.
Input sequence is randomly masked out before being fed into the GRU encoder and the feature-wise self-attention transformer.
Both the GRU enncoder output ($H$) and the transformer output ($C$) are then concatenated and simultaneously processed by the classifier for observed action classification.
The final state $h \in H$ of the GRU encoder and the max-pooled self-attention transformer output $c_{max}$ are also concatenated to obtain the initial hidden state of the the GRU decoder. The decoder then generates future actions by recursively decoding this information.
}
\label{fig.model}
\end{figure}

\subsection{Model description}
A visual illustration of our model is shown in figure~\ref{fig.model}.
Given the input feature sequence, first we perform feature masking to model temporal variations.
Next, we use two temporal encoding methods to capture two different types of temporal information.
The GRU encoder captures the temporal evolution of the signal at each time step while the "feature self-attention" captures the importance of each feature dimension in the input signal to better model the context information of each video.
Finally the observed features are classified to obtain actions of the observed data.
The GRU decoder processes temporal information as well as contextual information to generate actions of the future. Next we describe the details of each component.

Our model uses an encoder-decoder architecture, but with an additional classifier for predicting the actions of the observed part of the input, and a transformer to perform feature-wise self-attention on the input. 
The observed action classifier learns the semantics of observed activity of the person. This helps to better understand what the human is doing now and would help us when predicting what the human will do in the future.
The GRU encoder captures the temporal aspects of the input while our transformer captures relations between the feature dimensions in the feature space. The feature-wise transformer helps to better capture the contextual information.
The details for our approach are given in the rest of this section below.

\subsubsection{Feature extraction} 
Given the RGB video, we first use the I3D model \cite{Carreira2017} to extract spatio-temporal features from the video, which we denote by $ \mathbf{X} = \left< \mathbf{x}_{1}, \mathbf{x}_2, \cdots, \mathbf{x}_T \right >$ where $\mathbf{X} \in \mathbb{R}^{T \times d} $ where $d$ is the feature dimension and $T$ is the feature sequence length.

\subsubsection{Feature masking} 
There are large variations in daily human actions.
For example an activity such as "making a cup of tea" can be executed in many different ways. 
To capture the large variation with limited amount of training data, we propose a new video feature masking technique.
We mask $m$\% of the observed features of the video sequence by selecting frame features at random and set those feature values of those to zero, where $m$ is a hyperparameter. 
Feature masking completely removes information from a given feature, thus allowing us to augment the video features to be more robust to missing frames and missing action parts.
As there are large variations in the way an action can be performed, the feature masking helps to augment video data and allows us to capture temporal variations from training data.
Specifically, this is useful for longer videos and when the amount of training data is limited.
Masking has a different impact compared to drop-out. Drop-out randomly drops neuron connections in the network and avoids the model becoming overly reliant on certain neurons.
In contrast, masking on temporal data allows us to better capture temporal variations in the training signal and obtain a more robust model.

\subsubsection{GRU Encoder} 
The GRU encoder consists of a bi-directional GRU cell. 
We denote the GRU encoder hidden states by $\mathbf{H} = \left< \mathbf{h}_1, \mathbf{h}_2, \cdots, \mathbf{h}_T \right>$ with each vector of dimension $d_h$.
The initial hidden state of the GRU encoder $\mathbf{h}_0$ is set to zero.
The GRU encoder denoted by $f_e$ encodes the input $\mathbf{x}_t$ for time step $t$ as follows: 
\begin{equation}
\mathbf{h}_t = f_e(\mathbf{x}_t, \mathbf{h}_{t-1})
\end{equation}
where $\mathbf{h}_t$ is the hidden state at time step $t$.
The GRU is the temporal encoder of our method and it captures the temporal evolution of actions in the feature space.

\subsubsection{Feature-wise Transformer Encoder}
Contextual information is important for action understanding.
In this section we propose a new Transformer~\cite{vaswani2017attention}-based model to  identify the most relevant set of feature dimensions and then use that as an attention mechanism to obtain contextual video representation.
Traditionally, attention mechanism in most neural methods attempt to find salient feature vectors.
In contrast, we attempt to identify \emph{the most relevant feature dimensions} for action prediction using self-attention given a video.
We perform self-attention on the feature dimension of the input by using the scaled dot product attention as in \cite{vaswani2017attention} to obtain the self-attention output as follows:
\begin{equation}
\text{Attention}(\mathbf{Q},\mathbf{K},\mathbf{V}) = \text{softmax}\left(\frac{\mathbf{QK^\top}}{\sqrt{d_k}} \right)\mathbf{V}
\end{equation}
with matrices $\mathbf{Q}, \mathbf{K}, \mathbf{V} \in \mathbb{R}^{d \times d_{\text{model}}}$, $d_{\text{model}} = T$ and $d_k = T$.
One key difference is that the term $\mathbf{QK^\top}$ is a gram matrix of size $T \times T$ in traditional attention  whereas in our case it is a cross-correlation matrix of size $d \times d$. Softmax normalized cross-correlation is used to attend important feature dimensions of a given video.
We use multiple attention heads to allow the model to jointly attend to information from different representation subspaces. 
Using $h$ attention heads, we obtain the multi-headed attention output $\mathbf{A}$ as follows:
\begin{equation}
\mathbf{A} = \text{Concat}(\text{head}_1, \dotsc, \text{head}_h)W^O
\end{equation}
\\
where $\text{head}_i = \text{Attention}(\mathbf{Q'}W_i^Q,\mathbf{K'}W_i^K,\mathbf{V'}W_i^V)$ with matrices $\mathbf{Q'} = \mathbf{K'} = \mathbf{V'} = \mathbf{X}^\top$, projection matrices $W_i^Q, W_i^K, W_i^V \in \mathbb{R}^{d_\text{model} \times d_k}$ and $W^O \in \mathbb{R}^{hd_k \times d_\text{model}}$.
Each attention head is responsible for a subspace of transformed features and a corresponding cross-correlation attention matrix.
We then employ a residual connection followed by layer normalization to obtain
$\mathbf{B} = \text{LayerNorm} (\mathbf{X}^\top + \mathbf{A})$.
Finally we obtain the transformer encoder outputs $\mathbf{C}$ by using a feed forward layer:
\begin{equation}
\mathbf{C} = \text{LayerNorm}\left(\mathbf{B} + \mathbf{W}_{a_2} \times \text{ReLU} \left(\mathbf{W}_{a_1}\mathbf{B} \right)\right)^\top
\end{equation}
with projection matrices $\mathbf{W}_{a_1} \in \mathbb{R}^{d_\text{model} \times d_\text{ff}}$ and $\mathbf{W}_{a_2} \in \mathbb{R}^{d_\text{ff} \times d_\text{model}}$.
This Feature-wise Transformer Encoder output $\mathbf{C}$ is later temporally max pooled to obtain a contextualized video representation that is complementary to reccurent temporal information provided by the GRU Encoder.

\subsubsection{Observed Action Classifier}
In order to predict future actions accurately, a model should have a better semantic understanding of already performed actions.
A model that is able to accurately classify observed frames might be able to understand the action sequences better and therefore it will be able to predict future frames accurately.
However, it is important to have a good temporal representation of observed data.
Therefore, we makes use of two types of information extracted from the observed video sequence.
Given GRU encoder hidden states $\mathbf{H} = \left< \mathbf{h}_1, \mathbf{h}_2, \cdots, \mathbf{h}_T \right>$ and transformer encoder outputs $\mathbf{C} = \left< \mathbf{c}_1, \mathbf{c}_2, \cdots, \mathbf{c}_T \right>$, we concatenate both outputs to predict the observed actions. 
The GRU output is recurrent in nature and models the temporal features.
The transformer encoder outputs uses self-attention to obtain a temporal encoding that models the context of the action better.
Specifically, the features are given relative weights.
Observed frame classification is performed as an auxiliary task to regularize the prediction of future actions. The observed action score vector $\mathbf{\hat{y}}^{obs}_t$ at time step $t$ is given by: 
\begin{equation}
\mathbf{\hat{y}}^{obs}_t = \left( \mathbf{W}_\text{obs} \times (\mathbf{h}_t \oplus \mathbf{c}_t) \right)
\end{equation}
where $\oplus$ is the concatenation operation, $\mathbf{W}_\text{obs} \in \mathbb{R}^{2d_h \times N}$ is a learned parameter and $N$ is the number of action classes.
The above architecture makes use of both traditional recurrent GRU models and modern modified Transformer to capture video information to action classification.

\subsubsection{GRU Decoder}
The GRU decoder consists of a forward directional GRU cell and is used to prediction future actions. We denote the GRU decoder by $f_d$. The hidden state of the GRU decoder $\mathbf{h}^D_t$ and future action score vector $\mathbf{\hat{y}}^f_t$ at time step $t$ are given by the following equations:
\begin{equation}
\mathbf{h}^D_t = f_d(\mathbf{y}^f_{t-1} , \mathbf{h}^D_{t-1})
\end{equation}
\begin{equation}
\mathbf{\hat{y}}^f_t = \mathbf{W}_\text{fut} \times \mathbf{h}^D_t
\end{equation}
\\
with $\mathbf{W}_\text{fut} \in \mathbb{R}^{2d_h \times N}$ as a learned parameter, $\mathbf{h}^D_0 = \mathbf{h}_T \oplus \mathbf{c}_{max}$ where $\mathbf{c}_{max}$ is the temporally max-pooled vector obtained from the Transformer i.e. $\mathbf{c}_{max} = \max(\mathbf{c}_1, \mathbf{c}_2, \cdots, \mathbf{c}_T)$. Here $\mathbf{y}^D_0$ is set to SOS (start of sequence) symbol.
To summarize the transformer outputs, we use the max-pooling operator. This allows us to obtain a single context vector of the video and that representation is utilized along with the final GRU state to decode and obtain future predictions.

\subsubsection{Loss Function}
We use the cross entropy loss for training. The loss for predicting the observed actions is given by the cross entropy loss between the predicted $\mathbf{\hat{y}}^{obs}_t$ and the ground truth action $\mathbf{y}_t$ at time step $t$. Similarly, the loss for prediction future actions is given by the cross entropy loss between the predicted $\mathbf{\hat{y}}^{fut}_s$ and the ground truth action $\mathbf{y}_s$ at time step $s$. Denoting the cross entropy loss function by $L$, the combined loss $L_{total}$ is given by: 
\begin{equation}
\displaystyle L_{total} = \sum_{s=1}^{n_f} L(\mathbf{\hat{y}}^f_s, \mathbf{y}_s) + \beta \times \sum_{t=1}^T L(\mathbf{\hat{y}}^{obs}_t, \mathbf{y}_t)
\label{eq.loss}
\end{equation}
where $\beta$ is the regularization parameter.

\section{Experiments}
\subsection{Dataset}
We evaluate our model on two datasets: Breakfast \cite{Kuehne2014} and 50Salads \cite{Stein2013}.
Breakfast dataset \cite{Kuehne2014} contains 1,712 videos with 52 actors making breakfast dishes. There are 48 fine-grained actions classes and four splits. Each video consists of 6.8 actions per video on average. 
We use four splits as done in \cite{Kuehne2014} for training and testing.
50Salads \cite{Stein2013} dataset consists of 50 videos with 17 fine-grained action classes. The average length of a video is 6.4 minutes and the average action instances per video is 20.
We evaluate using the same five-fold cross validation method on the standard splits in the original paper.

\subsection{Experimental Setup}
For our experiments, we observe p\% of the video and predict actions for the next q\% of the video assuming  length  of  video is known and start and end times of each action are provided for the training videos.
The metrics used for evaluation is mean per class accuracy as done in \cite{AbuFarha2018} and \cite{ke2019time}.
Unless otherwise specified, we use effective I3D features \cite{Carreira2017} as the video representation for all datasets.
We first fine-tune the I3D network for video action classification  using the provided video level annotations. We then obtain a sequence of features where each feature is of dimension $d=1024$.

For our experiments, we use a batch size of 32 and learning rate of $10^{-4}$ with Adam optimizer. We randomly select 10\% of the training data to be used for validation and apply early stopping for our training, with a maximum of 100 training epochs for Breakfast dataset and 500 epochs for 50Salads dataset. We set the feature masking percentage m to be 10 for Breakfast dataset and 15 for 50Salads dataset.
We apply a dropout of 0.5 on the input of the GRU encoder and decoder. The GRU encoder hidden state dimension is set to 512. The embedding dimension for the output actions is 512.
We use only 1 transformer encoder layer with 5 attention heads and $d_\text{ff}$ set to 2048. The regularization parameter $\beta$ is set to 0.5.

\subsection{Architecture validation}
First, we evaluate the impact of each component in our architecture.
Table \ref{tbl.architecture} shows us the results of our model by varying different components to show the efficacy of each component in our architecture. The 5 columns under the Activation heading represents 5 different models comprising of the components where there is a check mark present. The last 2 rows shows the average mean per class accuracy taken over all 8 combinations of observation and prediction percentages for the Breakfast and 50Salads dataset respectively. Following \cite{AbuFarha2018} and \cite{ke2019time}, we set the observation percentages to 20\% and 30\% and prediction percentages to 10\%, 20\%, 30\% and 50\%.

Comparing the first two models, we can see that the feature-wise self attention component improves the average mean per class accuracy from 21.13\% to 21.86\% for the Breakfast dataset, and with a more significant improvement from 18.49\% to 22.54\% for the 50Salads dataset. This shows that the feature-wise self-attention has successfully identified relevant feature dimensions to aid in the action prediction.

The second and third model shows that adding the observed classifier gives a slight improvement over the results from 21.86\% to 22.03\% for the Breakfast dataset and from 22.54\% to 23.22\% for the 50Salads dataset. By having a semantic understanding of the observed actions, the model is able to better anticipate future unobserved actions. In contrast to prior models which relied on inferred action scores of observed data to forecast actions for the future, our model is less reliant on potentially erroneous observed classification results. We use these action classifiers as the auxiliary task using a multi-task objective. Later in the experiments, we analyze the influence of the observed classifier where we show how $\beta$ affects the results.

The results of fourth model with input feature masking compared with the third model without masking tells us that masking the input features during training can benefit the robustness of the model and improves the average mean over classes accuracy from 22.03\% to 22.39\% for the Breakfast dataset and from 23.22\% to 25.59\% for 50Salads dataset.

To further demonstrate the effectiveness of feature-wise self-attention, we conducted another experiment using the third model and replacing the feature-wise self-attention with temporal self-attention where the self-attention is performed over the temporal dimension. 
In fact, this alternative is the most common way self-attention is used. In this paper we use self-attention for an orthogonal task of finding the relevant feature dimensions.
The traditional way of using self-attention transformer results in a drop in performance from 22.03\% to 19.44\% for Breakfast dataset and from 23.22\% to 21.66\% for the 50Salads dataset. This shows the impact of our new feature-wise self attention along with the proposed new architecture that exploits both recurrent and self-attention-based models.

We conclude from this experiment that our new architecture for action forecasting is effective compared to considered alternative architectures. Specifically, the use of observed feature classifier, the masking technique and feature-based-self-attention module are complimentary and effective for action forecasting on two challenging datasets over $4 \times 8$\footnote{Breakfast dataset has 4 splits and 8 experimental configurations} and $5 \times 8$\footnote{50Salads dataset has 5 splits and 8 experimental configurations} repeated experiments.

\begin{table}[t!]
\centering
\resizebox{.99\columnwidth}{!}{
\begin{tabular}{|l|l|l|l|l|l|} \hline
Component & \multicolumn{5}{|c|}{Activation} \\ \hline
GRU Encoder-Decoder & \checkmark & \checkmark & \checkmark & \checkmark &  \checkmark \\ 
Feat Self Att. & & \checkmark & \checkmark & \checkmark & \\ 
Observed Classifier & & & \checkmark & \checkmark & \checkmark\\
Masking & & & & \checkmark & \\
Temporal Self Att. & & & & & \checkmark \\ \hline
Breakfast (mean over all \%) & 21.13 & 21.86 & 22.03 & \textbf{22.39} & 19.44\\
50Salads (mean over all \%)  & 18.49 & 22.54 &  23.22 & \textbf{25.59} & 21.66 \\ \hline
\end{tabular}
}
\caption{Evaluating the impact of architecture. We evaluate the impact of each component in our architecture. We report mean classification accuracy over all observation and prediction pecentages.}
\label{tbl.architecture}
\end{table}

\begin{figure}[t]
	\includegraphics[width=.49\columnwidth]{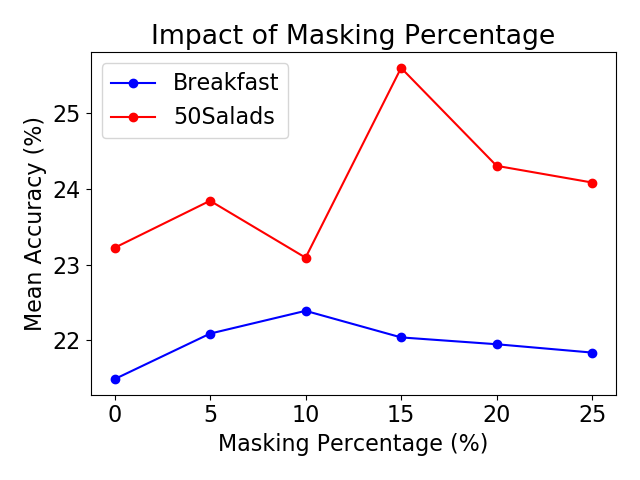}
	\includegraphics[width=.49\columnwidth]{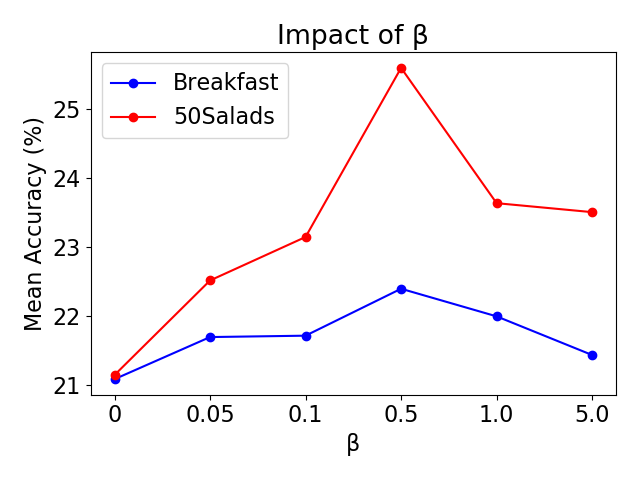}
	\caption{Left: Impact of masking percentage on both Breakfast and 50Salads datasets. Right: Impact of $\beta$ for observed classifier on both Breakfast and 50Salads datasets.}
\label{fig.masking}
\end{figure}
\begin{table*}[t]
\centering
\resizebox{.99\columnwidth}{!}{
\smallskip\begin{tabular}{|l|c|c|c|c|c|c|c|c|} \hline
observation (\%) & \multicolumn{4}{|c|}{20\%} & \multicolumn{4}{|c|}{30\%} \\ \hline
prediction  (\%) & 10\% & 20\% & 30\% & 50\%  & 10\% & 20\% & 30\% & 50\% \\ \hline
Grammar\cite{AbuFarha2018}    		& 16.60   & 14.95   & 13.47   & 13.42   & 21.10   & 18.18  & 17.46  & 16.30 \\ 
Nearest Neighbor\cite{AbuFarha2018} & 16.42   & 15.01   & 14.47   & 13.29   & 19.88   & 18.64  & 17.97  & 16.57 \\
RNN\cite{AbuFarha2018} 				& 18.11   & 17.20   & 15.94   & 15.81   & 21.64   & 20.02  & 19.73  & {19.21} \\ 
CNN\cite{AbuFarha2018}				& 17.90   & 16.35   & 15.37   & 14.54   & 22.44   & 20.12  & 19.69  & 18.76 \\ 
Time-Condition~\cite{ke2019time}    & 18.41   & 17.21   & 16.42   & 15.84   & 22.75   & 20.44& 19.64 & 19.75 \\ \hline
Ours   & \textbf{23.14} & \textbf{22.30} & \textbf{21.14} & \textbf{20.69} & \textbf{24.47} & \textbf{23.29} & \textbf{21.77} & \textbf{22.30} \\ \hline
\end{tabular}
}
\caption{Comparison of action forecasting methods on Breakfast dataset only using features. Mean per-class accuracy is reported.}
\label{tbl:mainbreakfast}
\end{table*}

\begin{table*}[t]
\centering
\resizebox{.99\columnwidth}{!}{
\begin{tabular}{|l|c|c|c|c|c|c|c|c|} \hline
observation (\%) & \multicolumn{4}{|c|}{20\%} & \multicolumn{4}{|c|}{30\%} \\ \hline
prediction  (\%) & 10\% & 20\% & 30\% & 50\%  & 10\% & 20\% & 30\% & 50\% \\ \hline
Grammar\cite{AbuFarha2018}    			& 24.73 & 22.34 & 19.76 & 12.74 & 29.65 & 19.18 & 15.17 & 13.14  \\ 
Nearest Neighbor\cite{AbuFarha2018}    	& 19.04 & 16.1  & 14.13 & 10.37 & 21.63 & 15.48 & 13.47 & 13.90  \\
RNN\cite{AbuFarha2018} 					& 30.06 & 25.43 & 18.74 & 13.49 & 30.77 & 17.19 & 14.79 & 9.77	 \\ 
CNN\cite{AbuFarha2018}				    & 21.24 & 19.03 & 15.98 & 9.87  & 29.14 & 20.14 & 17.46 & 10.86 \\ 
Time-Condition~\cite{ke2019time} 		& 32.51 & \textbf{27.61} & 21.26 & 15.99 & \textbf{35.12} & \textbf{27.05} & 22.05 & 15.59 \\ \hline
Ours   & \textbf{34.88} & 26.91 & \textbf{21.56} & \textbf{22.35} & 28.13 & 26.83 & \textbf{22.82} & \textbf{21.21} \\ \hline
\end{tabular}
}
\caption{Comparison of action forecasting methods on 50Salads dataset only using features. Mean per-class accuracy is reported.}
\label{tbl:pq.50salad}
\end{table*}


\subsection{Evaluating the impact of masking percentage}
In this section we evaluate the impact of masking percentage $m$ on action forecasting performance.
Figure \ref{fig.masking} (\textbf{left}) shows the mean accuracy over different masking percentages for both Breakfast and 50Salads datasets.

For the Breakfast dataset, we see an improvement in the mean accuracy with just 5\% feature masking from 21.49\% to 22.09\%, with a further improvement to 22.39\% when the masking percentage is increased to 10\%. However, further increasing the masking percentage impacts the results negatively. For the 50Salads, we see a similar trend as we introduce feature masking at 5\%, an improvement in mean accuracy follows. The results drops slightly when the percentage is increased to 10\%, but recovers and attains its best performance at 15\%. Further increase of the masking percentage results in a drop in the mean accuracy.

This shows that as we mask more features, the model is being trained to deal with missing features and thus the robustness of the model is improved. However, as the masking percentage is increased even further, we see a drop in the performance of the model. This demonstrates that if the masking percentage is too high, there will not be enough input features to make a proper prediction on future actions and thus is detrimental to the performance.

We also experimented with randomly shuffling certain percentage of the observed features to further model temporal variations in the input signal.
However, this strategy did not help us to improve results as the order of features are important in some activities. It appears that preserving the temporal order of observed input is important for prediction.
It seems that our model can recover some missing actions or the part of observed actions  when we use 5\%-15\% of masking.
This helps our model to become more invariant to temporal variations in the input sequence.
In fact, such a strategy interestingly increases the temporal robustness of the model for action forecasting.
By default in all our experiments we also use dropout with a dropout probability of 0.5. 
Indeed, dropout is complementary to temporal feature masking and in our experiments removal of dropout reduce the overall performance.

\subsection{Evaluating the impact of $\beta$ for observed classifier}
In this section we evaluate the impact of the observed action classifier by changing the value of $\beta$ in the overall loss function (Equation \ref{eq.loss}). In Equation ~\ref{eq.loss}, the $\beta$ weights the observed action loss.
Figure \ref{fig.masking} (right) shows the mean accuracy over different $\beta$ values for Breakfast and 50Salads datasets.   
For the Breakfast dataset, there is a steady increase in the mean accuracy from 21.08\% to 22.39\%  as the $\beta$ value is increased from 0 to 0.5. There is a similar trend in the 50Salads dataset where an increase in the $\beta$ value from 0 to 0.5 results in an increase in mean accuracy from 21.14\% to 25.59\%. For both datasets, further increase of the $\beta$ value results in a drop in performance. As the $\beta$ value is increased from 0.5 to 5, the mean accuracy drops from 22.39\% to 21.43\% for the Breakfast dataset and from 25.59\% to 23.50\% for the 50Salads dataset.
With a larger $\beta$ value, more weight is placed on the observed action loss and the model will place more emphasis on classifying the observed action during training. The results show that having and observed action classifier helps in model in predicting future actions more accurately as it has a better understanding of the semantic content of the observed video. However, if the $\beta$ value is too high, there is not enough emphasis placed on predicting future actions and the performance will be lower. Our result shows that having the observed action loss be about half of the future action loss is the most appropriate ratio.
\subsection{Comparison to state-of-the-art}
In this section we compare our results with the state-of-the-art action forecasting models presented in two prior works~\cite{AbuFarha2018,ke2019time}.
We report results for Breakfast dataset in Table~\ref{tbl:mainbreakfast} and 50Salads in Table~\ref{tbl:pq.50salad} respectively.
As before, we report mean per class accuracy over several observation and prediction percentages.
On Breakfast dataset we consistently outperform all prior models such as RNN and CNN models of \citeauthor{AbuFarha2018} and the time conditioned action forecasting model of~\citeauthor{ke2019time}.
It should be noted that the 50Salads has less variations compared to the Breakfast dataset.
In 50Salads dataset our model outperforms time-conditioned model in five cases.
Interestingly, our model outperforms the time-conditioned model when we observe less amount of data (20\%) and when predicting far into the future (30\% and 50\%).
Our results are excellent on Breakfast dataset and quite comparable on 50Salads dataset when compared with recent state-of-the-art models for action forecasting.

\section{Conclusion}
Action forecasting from videos is an important problem, yet not extensively investigated in the literature.
In this paper, we have presented a new architecture for action forecasting from videos by extending encoder-decoder architecture where the GRU encoder is complimented by feature-wise self-attention transformer model to identify relevant feature dimensions for the action forecasting task.
A combination of auxiliary observed action classifier and feature masking improve the results consistently.
We have analyzed the impact of each contribution of this paper and can conclude that feature-wise self-attention, auxiliary observed action classification and temporal feature masking is useful for action forecasting.
Careful analysis of our results shows that predictions are smooth and more frequent action transitions are predicted accurately (see supplementary material).
Perhaps feature-wise self-attention transformer is useful for other sequence to sequence tasks and our architecture that consists of temporal GRU and feature-wise  transformer can be used to solve related problems.

\bibliography{main.bib}

\end{document}